\documentclass[11pt,reqno]{amsart}
\usepackage{amsaddr}

\usepackage[margin=1in]{geometry}
\usepackage{fancyref}
\usepackage{bbm}

\usepackage{algorithm}
\usepackage[noend]{algpseudocode}

\usepackage[utf8]{inputenc} 
\usepackage[T1]{fontenc}    
\usepackage{hyperref}       
\usepackage{url}            
\usepackage{booktabs}       
\usepackage{amsfonts}       
\usepackage{nicefrac}       
\usepackage{microtype}      
\usepackage{xcolor}         
\usepackage{amsmath}
\usepackage[capitalize]{cleveref}
\usepackage{bm}
\usepackage{bbm}
\usepackage{mathtools}
\usepackage{amsthm}

\usepackage[T1]{fontenc}

\usepackage{microtype}
\usepackage{graphicx}
\usepackage{booktabs} 
\usepackage{natbib}
\usepackage{siunitx}
\usepackage{appendix}

\makeatletter
\def\paragraph{\@startsection{paragraph}{4}%
  \z@\z@{-\fontdimen2\font}%
  {\normalfont\bfseries}}
\makeatother

\usepackage{amsmath,amsfonts,bm}

\def\1{\bm{1}}

\DeclareMathAlphabet{\mathsfit}{\encodingdefault}{\sfdefault}{m}{sl}
\SetMathAlphabet{\mathsfit}{bold}{\encodingdefault}{\sfdefault}{bx}{n}

\usepackage{caption}
\usepackage{booktabs}
\usepackage{multirow}
\usepackage{enumitem}

\usepackage{subcaption}
\usepackage{hyperref}
\usepackage{url}
\usepackage{amsmath,amsthm}
\usepackage{amssymb,bm}
\usepackage{bbm}
\usepackage{makecell}
\newtheorem{theorem}{Theorem}

\newtheorem{lemma}{Lemma}

\newtheorem{corollary}{Corollary}
\usepackage{pgfplots}
\pgfplotsset{compat=1.10}
\usetikzlibrary{
        patterns,
        pgfplots.fillbetween,
    }

       \tikzset{
        hatch distance/.store in=\hatchdistance,
        hatch distance=10pt,
        hatch thickness/.store in=\hatchthickness,
        hatch thickness=2pt,
        hatch color/.store in=\hatchcolor,      
        hatch color=black,                      
    }
    \pgfdeclarepatternformonly[%
        \hatchdistance,%
        \hatchthickness,%
        \hatchcolor
            ]{flexible hatch}
    {\pgfqpoint{0pt}{0pt}}
    {\pgfqpoint{\hatchdistance}{\hatchdistance}}
    {\pgfpoint{\hatchdistance-1pt}{\hatchdistance-1pt}}%
    {
        \pgfsetlinewidth{\hatchthickness}
        \pgfpathmoveto{\pgfqpoint{0pt}{0pt}}
        \pgfpathlineto{\pgfqpoint{\hatchdistance}{\hatchdistance}}
        \pgfsetstrokecolor{\hatchcolor}
        \pgfusepath{stroke}
    }
\providecommand{\nor}[1]{\ensuremath{\left\lVert {#1} \right\rVert}}

\usepackage[makeroom]{cancel}

\hypersetup{
     colorlinks = true,
     linkcolor = red,
     anchorcolor = blue,
     citecolor = teal,
     filecolor = blue,
     urlcolor = black
     }
 \usepackage{tcolorbox}

\newtcolorbox{assbox}{colback=black!5!white,colframe=black!75!black}
  \newtcolorbox{thmbox}{colback=blue!5!white,colframe=black!75!black}

\title[Revisiting Group Relative Policy Optimization]{Revisiting Group Relative Policy Optimization:\\ Insights into On-Policy and Off-Policy Training}
\author[]{Youssef Mroueh$^{\star}$, Nicolas Dupuis$^{\dagger}$, Brian Belgodere$^{\star}$,\\ Apoorva Nitsure$^{\star}$, Mattia Rigotti$^{\star}$, Kristjan Greenewald$^{\star,\circ}$,\\ Jiri Navratil$^{\star}$ , Jerret Ross$^{\star}$,  Jesus Rios$^{\star}$ }
\address[Y.Mroueh]{$\star$ IBM Research, $\dagger$ IBM Quantum, $\circ$ MIT-IBM Watson Lab}

\definecolor{darkgreen}{rgb}{0.0, 0.5, 0.0}
\definecolor{darkred}{rgb}{0.5, 0.0, 0.0}

\definecolor{dorkgreen}{rgb}{0.0, 0.8, 0.0}

 \usepackage{tcolorbox}

\begin{document}

\maketitle

\begin{abstract}
We revisit Group Relative Policy Optimization (GRPO) in both on-policy and off-policy optimization regimes. Our motivation comes from recent work on off-policy Proximal Policy Optimization (PPO), which improves training stability, sampling efficiency, and memory usage. In addition, a recent  analysis of GRPO suggests that estimating the advantage function  with off-policy samples could be beneficial. Building on these observations, we adapt GRPO to the off-policy setting.
 We show that both on-policy and off-policy GRPO objectives yield an improvement in the reward. This result motivates the use of clipped surrogate objectives in the off-policy version of GRPO. We then compare the empirical performance of reinforcement learning with verifiable rewards in post-training using both GRPO variants. Our results show that off-policy GRPO either significantly outperforms or performs on par with its on-policy counterpart.
\end{abstract}

\section{Introduction}

Proximal Policy Optimization (PPO) \citep{schulman2015trust,schulman2017proximal} is a widely used algorithm in reinforcement learning. Reinforcement learning from Human Feedback \citep{NIPS2017_d5e2c0ad,stiennon2020learning,ouyang2022training,bai2022training} and Reinforcement Learning from Verifiable Rewards \citep{lambert2024t,shao2024deepseekmath}   are corner stones in post-training of large language models to align their preferences with human values and to enable reasoning and coding capabilities using verifiable rewards. \\

Group Relative Policy Optimization introduced in \citep{shao2024deepseekmath} alleviate the need of training a critic network in PPO and uses Monte-Carlo samples referred to as ``a group'' to estimate the advantage function via a standardized reward, where the mean and standard deviation statistics are estimated using the group. GRPO was used to train the Deepseek R1 reasoning models \citep{guo2025deepseekR1} and was adopted by the open-source community as a method of choice for post-training of large language models, with open-source implementations in several librarires such as TRL of HuggingFace \citep{vonwerra2022trl} and VERL \citep{deepscaler2025}.\\

Several recent works analyzed the loss implemented in GRPO such as \cite{vojnovic2025alignmentobjectivegrpo,mroueh2025reinforcementlearningverifiablerewards}.
The study in \cite{mroueh2025reinforcementlearningverifiablerewards} suggests that the iterative GRPO of \cite{shao2024deepseekmath} with sample reuse (i.e.\ for $\mu>1$ in \cite{shao2024deepseekmath}) leads to an off-policy estimation of the advantage and to a success rate amplification when using verifiable rewards.
Indeed, it has been observed empirically that this off-policy advantage estimation leads to an improved performance \citep{huggingface_openr1_update3}.\\

Motivated by these observations and the rich literature on off-policy PPO and RL, like work by \cite{queeney2021generalized,meng2023off,gan2024transductive,fakoor2020p3o} to cite a few
(see related work Section \ref{sec:relatedwork} for a larger account on this), in this paper we explore the extension of GRPO to the off-policy regime where the advantage is estimated using statistics coming from a different policy than the current policy.\\  

The main contributions of this paper are:
\begin{itemize}
\item We review in Section \ref{sec:op-GRPO} the iterative GRPO algorithm proposed in \cite{shao2024deepseekmath} and introduce in Section \ref{sec:off-GRPO} the off-policy GRPO. 
\item We show in Section \ref{sec:off-GRPO} that the  on-policy and off-policy advantages provides a lower bound on the policy improvement of the expected reward (Theorem \ref{lem:offpolicyimprovement} and Corollary \ref{lem:onpolicyimprovement}).
\item We state conditions under which optimizing the advantage leads to  improvements in the off-policy regime, namely, given that the off-policy stays in the vicinity of the current policy and the variance of the reward under the off-policy is non zero, maximizing the regularized off-policy advantage leads to  policy improvement. The regularization ensures that the updated policy stays close to the off-policy. 
\item Finally, armed with these results, we state the constrained policy optimization problem for off-policy GRPO in Section \ref{sec:clipped} and derive a clipped surrogate similar to the ones in off-policy PPO \citep{gan2024transductive} and obtain on-policy GRPO clipped objective as a particular case. 
\item We validate experimentally that training LLMs with off-policy GRPO leads to either improved or on par performance while potentially reducing the communication burden in serving the model in each iteration for inference. 
\end{itemize}

\section{On-Policy GRPO}\label{sec:op-GRPO}
Let $\mathcal{X}$ be the space of inputs (prompts in the context of LLMs) and $\mathcal{Y}$ the space of responses. We denote by $\rho_{\mathcal{X}}$ the distribution on inputs. 
We refer to the policy we want to optimize as $\pi(\cdot|x)$, which is a distribution on $\mathcal{Y}$ conditioned on $x\sim \rho_{\mathcal{X}}$.
For $k\geq 0$, let $\pi_{k}$ be the policy at the current step $k$. \\

The Group Relative Policy Optimization (GRPO) Clipped objective introduced in \cite{shao2024deepseekmath} is a variant of Proximal Policy Optimization (PPO) \citep{schulman2017proximal,schulman2015trust}, where the advantage is computed as a standardized reward function with mean and variances computed with respect to a group or Monte-Carlo samples of  size $G$ sampled from the current policy $\pi_{k}(.|x)$ for each $x$ independently.
For $\epsilon,\beta>0$ and given a reference policy $\pi_{\mathrm{ref}}$, the clipped objective optimization in  GRPO is defined as follows:
\[ \max_{\pi} \mathbb{E}_{y \sim \pi_{k}(\cdot|x)}\min\left(\frac{\pi(y|x)}{\pi_{k}(y|x)} A_{\pi_{k}}(x,y), ~\text{clip}\left(\frac{\pi(y|x)}{\pi_{k}(y|x)},1-\epsilon, 1+ \epsilon\right) A_{\pi_{k}}(x,y)\right) -\beta \mathsf{KL}(\pi||\pi_{\mathrm{ref}}), \]
where $\mathsf{KL}$ is the Kullback-Leibler divergence, and $A_{\pi_{k}}$ is the GRPO advantage function:
\[ A_{\pi_{k}}(x,y)= \frac{r(x,y)- \mathbb{E}_{\pi_k}r(x,y)}{\sqrt{\mathbb{E}_{\pi_k} (r(x,y)-\mathbb{E}_{\pi_k}r(x,y) )^2+ \varepsilon }}.\]
The advantage can be estimated from samples on ``a group'' of size $G$ for each $x$, we sample $y_1,\ldots,y_{G} \sim \pi_k(\cdot|x)$ and compute $r_{\ell}=r(x,y_{\ell}),$ $\ell=1,\ldots,G$. We refer to the group of reward conditioned on $x$ as $\{r_{\ell}\}$ and the estimated GRPO advantage is therefore \citep{shao2024deepseekmath}:
\[ \hat{A}_{\pi_{k}}(x,y_i) = \frac{r_i -\texttt{mean}(\{r_{\ell}\})}{\sqrt{\texttt{std}^2(\{r_{\ell}\})+ \varepsilon}}, \]
where \texttt{mean} and \texttt{std} are empirical mean and standard deviation respectively.
The statistics used to normalize the reward leading to the advantage function are estimated using the current policy $\pi_{k}$, and hence we refer to $A_{\pi_{k}}$ as the \emph{on-policy advantage}.\\

When compared with PPO, GRPO alleviates the need of training a critic network to compute the advantage and relies instead on standarized rewards that can be estimated efficiently using efficient inference frameworks such as vLLM \citep{kwon2023efficient} in the context of large language models.\\

\paragraph{GRPO with Verifiable Rewards and Success Rate Amplification} The iterative GRPO \citep{shao2024deepseekmath} has two  overlooked features: 
\begin{itemize}
  \item The algorithm suggests to optimize the policy $\pi$ for $\mu$ iterations fixing the samples from $\pi_{k}$, which inherently leads to an off-policy estimation of the advantage. 
  \item The algorithm suggests to do the training in stages while changing $\pi_{\mathrm{ref}}$ to the latest optimized policy with GRPO.
\end{itemize}

A recent analysis of GRPO with verifiable rewards, i.e.\ with binary rewards \citep{mroueh2025reinforcementlearningverifiablerewards}, suggests that this aforementioned  off-policy advantage estimation in \cite{shao2024deepseekmath} leads to  an implicit fixed point iteration that guarantees that the success rate of the GRPO-optimized policy is higher than the one of the reference policy.
This also explains the multi-stage nature of the iterative GRPO that changes the reference along the training iterations. 

Motivated by these observations, we propose to take a step back and analyze on-policy and off-policy GRPO.
In practice, in our proposed off-policy GRPO instead of just fixing the samples for $\mu$ iterations from $\pi_{k}$ as suggested in \cite{shao2024deepseekmath}, we use the policy $\pi_{k-\mu}$ to estimate the advantage for $\mu$ iterations with fresh samples in each iteration, and we refer to this as \emph{off-policy} advantage.   
 
\section{Off-Policy and On-Policy GRPO Reward Improvement}\label{sec:off-GRPO}

We introduce in this Section off-policy GRPO, and analyze conditions under which policy reward improvement is possible in both the on-policy and off-policy regimes.
Towards that goal we start by some preliminary definitions.

Define the expected reward of a policy given $x\sim \rho_{\mathcal{X}}$:
\begin{equation}
J(\pi (\cdot| x)) = \mathbb{E}_{y \sim \pi(\cdot|x) } r(x,y) 
\label{eq:expReward}
\end{equation}

For $k\geq 0$, let $\pi_{k}$ be the policy at the current step $k$ and  $\alpha(\cdot|x)$ be  a policy used for off-policy sampling, where typically we consider $\alpha(\cdot|x) = \pi_{k-v}(\cdot | x)$, for $0\leq v <k$. \footnote{Note in Section \ref{sec:op-GRPO} we referred to this as $\pi_{k-\mu}$ so we keep close to notation used in the original GRPO paper. We will use $v$ instead of $\mu$ in the rest of the paper.}

Define the  mean and standard deviation of the off-policy reward, i.e.\ under policy $\alpha$:
 \[{\mu_{\alpha,r} (x) =  \mathbb{E}_{y \sim \alpha(\cdot|x)} r(x,y)}\]
and 
 \[\sigma_{\alpha,r}(x) = \sqrt{ \mathbb{E}_{y \sim \alpha(\cdot|x)}( r(x,y) - \mu_{\alpha,r} (x))^2 },\]
and denote for ${0<\varepsilon<1}$:
\[\sigma_{\alpha,r,\varepsilon}(x) =\sqrt{\sigma^2_{\alpha,r}(x) + \varepsilon}.\]



The GRPO advantage function computed using the off-policy distribution $\alpha$ is defined as the whitened reward, as follows: 
\begin{equation}
 A_{\alpha} (x,y)   =  \frac{r(x,y) - \mu_{\alpha,r} (x)  }{\sigma_{\alpha,r,\varepsilon}(x)}.
 \label{eq:advantage general}
 \end{equation}

Our goal is to maximize the expected advantage function using importance sampling under the policy $\alpha$:
\begin{equation}
\mathcal{L}_{\alpha}(\pi (\cdot | x)) = \mathbb{E}_{y \sim \alpha(\cdot|x)} \frac{\pi(y|x)}{\alpha(y|x)} A_{\alpha}(x,y) 
\label{eq:expAdvantage}
\end{equation}

If $\alpha= \pi_{k}$, we obtain the online policy objective function of GRPO, where the advantage is computed with the current policy $\pi_{k}$, i.e.\ using $A_{\pi_{k}}(x,y)$.

\subsection{Policy Improvement in GRPO}

Note that our goal is to optimize the expected reward under $\pi$, $J(\pi)$ given in eq.\ \eqref{eq:expReward}, but instead we use the expected advantage $\mathcal{L}_{\alpha}(\pi)$ -- where the advantage is computed using $\alpha$ -- given in eq.\ \eqref{eq:expAdvantage}.
Hence, our goal in what follows is to provide a lower bound on $J(\pi(\cdot|x )) - J( \pi_{k}(\cdot|x) )$ that involves $\mathcal{L}_{\alpha}(\pi)$, which guarantees that maximizing the expected advantage function leads to improvement in terms of expected rewards on the current policy $\pi_{k}$. 

Our lower bounds are given in Theorem \ref{lem:offpolicyimprovement} and  Corollary \ref{lem:onpolicyimprovement} and they involve the total variation distance $\mathbb{TV}$ defined as follows:
\[ \mathbb{TV}(m_1,m_2) = \frac{1}{2} \int |m_1-m_2|. \]

\begin{theorem} [Policy Improvement Lower Bound in \textbf{Off-Policy GRPO}] Assume that the reward is positive and bounded in $0\leq r \leq 1$. Let $\alpha$  be the off-policy distribution and $\pi_{k}$ the current policy. Then for any policy $\pi$ we have for all $x$ ($\rho_{\mathcal{X}}$ a.s.):\\

\boxed{J(\pi(\cdot|x )) - J( \pi_{k}(\cdot|x) ) \geq \mathcal{L}_{\alpha}(\pi (\cdot | x)) - 2~\frac{1- \sigma_{\alpha,r,\varepsilon}(x)}{\sigma_{\alpha,r,\varepsilon}(x)} ~\mathbb{TV}(\pi(\cdot|x),\alpha(\cdot|x)) - 2~\mathbb{TV}(\pi_k(\cdot|x),\alpha(\cdot|x))}
\label{lem:offpolicyimprovement}
\end{theorem}

If the reward is not bounded by $1$ we can scale it by $\nor{r}_{\infty}$ so it becomes in $[0,1]$, without this impacting the overall optimization problem. Note that this condition on the reward ensures that $\sigma_{\alpha,r}(x) \leq 1$ which is needed in the GRPO case to get the policy improvement lower bound. Indeed for bounded random variable in $[a,b]$ the variance is bounded by $\frac{(b-a)^2}{4}$, and hence we have $\sigma_{\alpha,r}(x) \leq \frac{1}{4}$, which guarantees that the term $\frac{1- \sigma_{\alpha,r,\varepsilon}(x)}{\sigma_{\alpha,r,\varepsilon}(x)} \geq 0$. 

For on-policy GRPO i.e.\ setting $\alpha=\pi_{k}$
in Theorem \ref{lem:offpolicyimprovement} we have the following corollary: 

\begin{corollary}
[Policy Improvement Lower Bound in \textbf{On-Policy GRPO}]
 Assume that the reward is positive and bounded, $0\leq r \leq 1$. Let  $\pi_{k}$ be the current policy, then for any policy $\pi$ we have for all $x$ ($\rho_{\mathcal{X}}$ a.s.):
\begin{center}
  \boxed{J(\pi(\cdot|x )) - J( \pi_{k}(\cdot|x) ) \geq \mathcal{L}_{\pi_{k}}(\pi (\cdot | x)) - 2 ~\frac{1- \sigma_{\pi_{k},r,\varepsilon}(x)}{\sigma_{\pi_{k},r,\varepsilon}(x)} ~\mathbb{TV}(\pi(\cdot|x),\pi_{k}(\cdot|x))}   
\end{center}
\label{lem:onpolicyimprovement}
\end{corollary}

Define : 
\[ M_{\alpha,r,\varepsilon}=\sqrt{\mathbb{E}_{x\sim \rho_{\mathcal{X}}}\frac{(1- \sigma_{\alpha,r,\varepsilon}(x))^2}{\sigma^2_{\alpha,r,\varepsilon}(x)} } \]
Integrating Theorem \ref{lem:offpolicyimprovement}  on $x$ (prompts) and applying Cauchy-Schwarz inequality we obtain: 
\begin{assbox}
\begin{align}
&\mathbb{E}_{x\sim \rho_{\mathcal{X}}}J(\pi(\cdot|x )) - \mathbb{E}_{x\sim \rho_{\mathcal{X}}}J( \pi_{k}(\cdot|x) )
 \geq \mathbb{E}_{x\sim \rho_{\mathcal{X}}}\mathcal{L}_{\alpha}(\pi (\cdot | x))\nonumber ..\\ 
 & ..- 2 M_{\alpha,r,\varepsilon}  (\mathbb{E}_{x\sim \rho_{\mathcal{X}}}\mathbb{TV}^2(\pi(\cdot|x),\alpha(\cdot|x)))^{\frac{1}{2}}
- 2\mathbb{E}_{x\sim \rho_{\mathcal{X}}} \mathbb{TV}(\pi_k(\cdot|x),\alpha(\cdot|x))\label{eq:IntegralLowerBound}
\end{align}
\end{assbox}

\paragraph{Interpreting the lower bound}
When compared with lower bounds for policy improvement in PPO (Theorem 1 in TRPO \citep{schulman2015trust}) and for off-policy PPO (Lemma 3.1 in transductive PPO \citep{gan2024transductive} and  Theorem 1 in Generalized PPO  \citep{queeney2021generalized}), we observe similar lower bounds with a crucial difference that the  constants weighting total variations are absolute constants for PPO whereas they are policy and data dependent for GRPO.
In particular, the dependency of the lower bound on: 
\[ \frac{1- \sigma_{\alpha,r,\varepsilon}(x)}{\sigma_{\alpha,r,\varepsilon}(x)} \]
is of interest. We can examine this quantity for verifiable rewards, for each $x$ the verifiable reward is a Bernouilli random variable with parameter $p$ the probability of success of the policy given $x$ \citep{mroueh2025reinforcementlearningverifiablerewards}. Hence we have: 
\[  \frac{1- \sigma_{\alpha,r,\varepsilon}(x)}{\sigma_{\alpha,r,\varepsilon}(x)} = \frac{1- \sqrt{p(1-p)+\varepsilon}}{\sqrt{p(1-p)+\varepsilon}} \]
Plotting this quantity as function of $p$ below, we observe that it diverges for fully correct and incorrect answers and this can indeed hurt the lower bound, as the negative terms in the lower bound will be dominating. It was suggested in DAPO \citep{yu2025dapoopensourcellmreinforcement} to filter out prompts with fully correct or incorrect answers, this will have the effect of controlling this term in the lower bound and keep that quantity bounded away from infinity. 

\begin{figure}[!ht]  
    \centering  
    \includegraphics[width=0.6\textwidth]{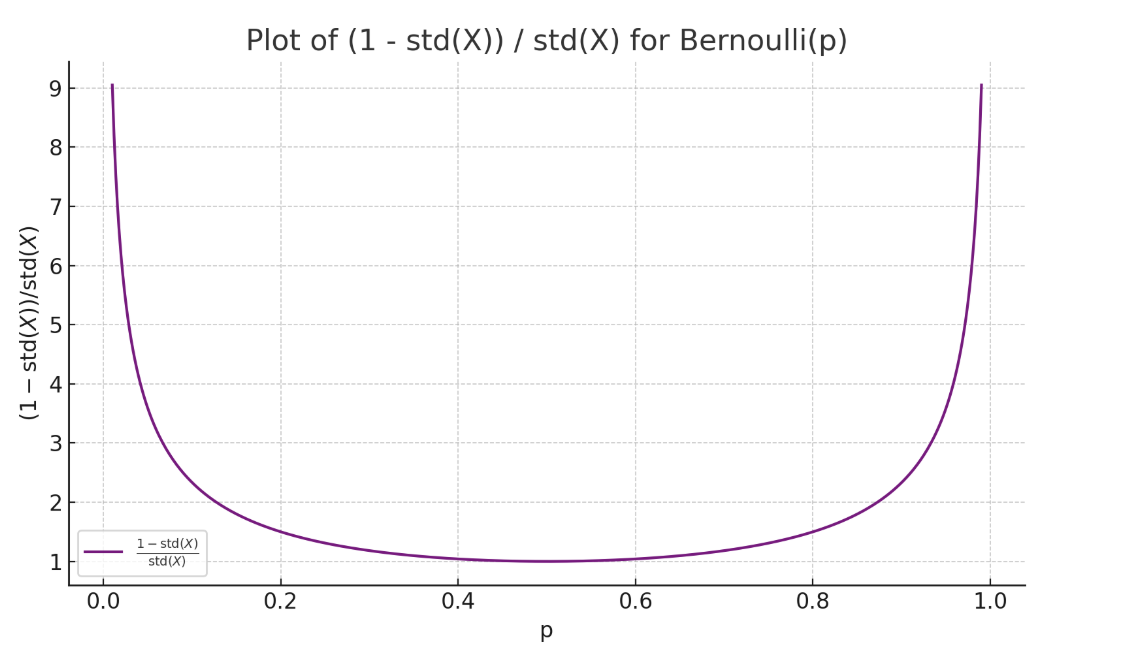}  
    \caption{ $\frac{1- \sigma_{\alpha,r,\varepsilon}(x)}{\sigma_{\alpha,r,\varepsilon}(x)}$  explodes when variance is zero, meaning for fully correct or wrong policies, this term dominates the lower bound. }  
    \label{fig:example}  
\end{figure}

\subsection{GRPO: From Constrained Optimization to Clipped Surrogate Objectives}\label{sec:clipped}
~~\\
\paragraph{From Penalized to $\mathsf{KL}$ Constrained Optimization} To maximize the lower bound in eq.\eqref{eq:IntegralLowerBound},   we see that the off-policy $\alpha$ needs to be in the vicinity of the current policy $\pi_{k}$, i.e.\ for $\mathbb{TV}(\alpha,\pi_{k})\leq \delta$  and that \boxed{M_{\alpha,r,0}<\infty} (variance terms not exploding). Under these assumptions, we can solve the following penalized problem :
 \[ \max_{\pi} \mathbb{E}_{x\sim \rho_{\mathcal{X}}} \mathcal{L}_{\alpha}(\pi (\cdot | x)) - 2~M_{\alpha,r,\varepsilon}  \sqrt{\mathbb{E}_{x\sim \rho_{\mathcal{X}}} \mathbb{TV}^2(\pi(\cdot|x),\alpha(\cdot|x))} .  \]

By virtue of Theorem \ref{lem:offpolicyimprovement}, maximizing this objective above leads to policy reward improvement. 

We can write this as a constrained optimization, there exists   $\Delta >0$ such that the following constrained optimization problem is equivalent:
\[ \max_{\pi} \mathbb{E}_{x\sim \rho_{\mathcal{X}}} \mathcal{L}_{\alpha}(\pi (\cdot | x)) \text{ subject to }   \mathbb{E}_{x\sim \rho_{\mathcal{X}}} \mathbb{TV}^2(\pi(\cdot|x),\alpha(\cdot|x)) \leq \Delta^2.  \]

By Pinsker inequality for two measures $m_1,m_2$ we have $\mathbb{TV}(m_1,m_2)\leq  \sqrt{\frac{1}{2} \mathsf{KL}(m_1,m_2)}$ and hence we can bound instead the $\mathsf{KL}$ divergence as follows:
\begin{equation}
\boxed{\max_{\pi} \mathbb{E}_{x\sim \rho_{\mathcal{X}}} \mathcal{L}_{\alpha}(\pi (\cdot | x)) \text{ subject to }  \frac{1}{2}~\mathbb{E}_{x\sim \rho_{\mathcal{X}}} \mathsf{KL}(\pi(\cdot|x),\alpha(\cdot|x)) \leq \Delta^2.}
\label{eq:KLConstrained}
\end{equation}

\paragraph{From Constrained Optimization to Clipped Surrogate Objectives} The objective in \eqref{eq:KLConstrained} is the same as in the original constrained PPO formulation \citep{schulman2015trust} with two key differences: the advantage is the whitened reward of GRPO where the statistics are computed using the off-policy $\alpha$ , and the advantage objective is computed using importance sampling from the off-policy $\alpha$, instead of $\pi_{k}$ in both cases. This is indeed related to objectives in off-policy PPO \citep{queeney2021generalized, gan2024transductive}.
A practical implementation of these objectives is through clipped surrogates \citep{schulman2015trust}.  

For $\epsilon \in [0,1]$ following \cite{gan2024transductive, queeney2021generalized} let us define:
\[f_{\epsilon}(r,r',a) = \min(r a, ~\text{clip}(r,\max(r'-\epsilon,0), r'+ \epsilon) ~a ).\]

The clipped off-policy GRPO objective for $\alpha $ such that \boxed{\mathbb{TV}(\alpha,\pi_{k})\leq \delta} and \boxed{M_{\alpha,r,0}<\infty} is therefore defined as follows :

\begin{equation}
\mathcal{L}^c_{\alpha}(\pi (\cdot | x)) = \mathbb{E}_{y \sim \alpha(\cdot|x)} f_{\epsilon}\left( \frac{\pi(y|x)}{\alpha(y|x)},\frac{\pi_k(y|x)}{\alpha(y|x)}, A_{\alpha}(x,y) \right)
\label{eq:expAdvantageclipped}
\end{equation}

Let us unpack this, we have: $f_{\epsilon}\left( \frac{\pi(y|x)}{\alpha(y|x)},\frac{\pi_k(y|x)}{\alpha(y|x)}, A_{\alpha}(x,y) \right) = ..$
\[
..=
\begin{cases}
  A_{\alpha}(x,y) \min\left(\frac{\pi (y| x )}{\alpha(y| x) }  , \frac{\pi_{k}(y|x)}{\alpha(y|x)}+\epsilon\right), & r(x,y) \geq \mu_{\alpha,r}(x) \\
A_{\alpha}(x,y) \max\left(\frac{\pi (y| x )}{\alpha(y| x) }  , \max(\frac{\pi_{k}(y|x)}{\alpha(y|x)}-\epsilon,0)\right), & r(x,y) < \mu_{\alpha,r}(x).
\end{cases}
\]
The clipping ensures that the ratio $\frac{\pi}{\alpha}$ remains bounded and is a relaxation of the $\mathsf{KL}$ (or the total variation distance). Since $\alpha$ needs to satisfy closeness to $\pi_{k}$ in order to ensure  improvement, the clipping objective incentivizes the difference between $\frac{\pi}{\alpha}-\frac{\pi_k}{\alpha}$ to not exceed $\epsilon$ \citep{gan2024transductive}.

In practice, the off-policy is $\alpha = \pi_{k-v}$ for a small $v \in [0,k)$. Given a small learning rate and a small $v$, the assumption that the policy $\pi_{k-v}$ doesn't deviate from $\pi_{k}$ is reasonable, and for $v$ small we can approximate $\frac{\pi_k}{\pi_{k-v}}$ by $1$.
We use this approximation in practice as we found it more stable, and given that this approximation is in practice used in off-Policy PPO (with sample reuse) as discussed in \cite{gan2024transductive} (See Section 4.1 in \cite{gan2024transductive}). \\

\paragraph{Back to On-Policy GRPO Clipped Objective} For $\alpha=\pi_{k}$, we obtain the clipped objective for on-policy GRPO \citep{shao2024deepseekmath}:
\begin{align*}
\mathcal{L}^c_{\pi_{k}}(\pi (\cdot | x)) &= \mathbb{E}_{y \sim \pi_{k}(\cdot|x)} f_{\epsilon}\left( \frac{\pi(y|x)}{\pi_k(y|x)},1, A_{\pi_{k}}(x,y) \right)\\
&= \mathbb{E}_{y \sim \pi_{k}(\cdot|x)}\min\left(\frac{\pi(y|x)}{\pi_{k}(y|x)} A_{\pi_{k}}(x,y), \text{clip}\left(\frac{\pi(y|x)}{\pi_{k}(y|x)},1-\epsilon, 1+ \epsilon\right) A_{\pi_{k}}(x,y)\right).
\label{eq:expAdvantageclippedonpolicy}
\end{align*}
~~\\

\begin{table}[ht!]
\centering
\begin{tabular}{|l|c|c|}
\hline
Method name & \textcolor{blue}{ \textbf{Update by fixed batch $i$}} & \textcolor{blue}{ \textbf{Update of Policy on Server $v$}}  \\
\hline
On-Policy GRPO \citep{shao2024deepseekmath}& $i=1$ &$v=1$ \\
\hline
Off-policy GRPO \citep{shao2024deepseekmath}&$i>1$ & $v=1$\\
\hline
Off-policy GRPO (this work) &$i=1$ & $v>1$  \\
\hline
\end{tabular}
\hfill
\caption{Training configurations in alg. \ref{alg:iter-grpo}: \texttt{(v1-i1)} is on-policy GRPO and \texttt{(v1-i10)} is an example of off-policy GRPO in \citep{shao2024deepseekmath}. Our off-policy GRPO corresponds e.g. to \texttt{(v10-i1)}.}
\label{tab:configs}
\vskip -0.2in
\end{table}

 \begin{algorithm}[ht!]
  \small
  \caption{Iterative GRPO with verifiable rewards, modified from  \cite{shao2024deepseekmath}}

  \begin{algorithmic}[1]
\State  \textbf{Input} initial policy model $\pi_{\theta_{\text{init}}}$; verifiable reward $r$; task prompts $\mathcal{D}$; 
\State \textbf{Hyperparameters }$\epsilon$, $\beta$, $S$,
\State \textcolor{blue}{$(i,v)$=(Number of SGD iteration by fixed batch,  Model update on vLLM server)} 
    \State \textbf{Policy model} $\pi_\theta \leftarrow \pi_{\theta_{\text{init}}}$ $\pi_{\mathrm{ref}}\gets \pi_{\theta_{\text{init}}}$
      \For {$s=1,\dots, S$}
      \For{$k = 1, \dots, M$}
      \State Sample a batch $\mathcal{D}_b$ from $\rho_{\mathcal{X}}$
      \If{\textcolor{blue}{\textbf{$k \bmod v = 0$}} }
      \State Update the old policy model on the vLLM server $\pi_{\theta_{\textrm{old}}} \leftarrow \pi_{\theta}$ 
      \EndIf
      \State Sample $G$ outputs $\{y_i\}_{i=1}^G \sim \pi_{\theta_{\textrm{old}}} (\cdot \mid x_i) $ for each question $x \in \mathcal{D}_b$
      \State Compute rewards $\{r_i\}_{i=1}^{G}$ for each sampled output $y_i$ by running verifiable reward $r$ 
      \State \textcolor{blue}{$\alpha \gets \pi_{\theta_{\textrm{old}}}$}
      \State Compute \textcolor{blue}{$A_{\alpha}(x,y_i)$}  using Equation \eqref{eq:advantage general}
      \For{\textcolor{blue}{GRPO iteration = 1, \dots, \textbf{$i$}}}  \Comment{\textcolor{blue}{$i$} is referred to as \textcolor{blue}{$\mu$} in Original GRPO}
        \State Update the policy model $\pi_{\theta}$ by maximizing the GRPO objective \eqref{eq:KLLC} with gradient ascent
       
      \EndFor
    \EndFor 
    \State $\pi_{\mathrm{ref}}\gets \pi_{\theta} $ \Comment{Swap reference with the latest model}
    \EndFor
    \State \noindent   \textbf{Output} $\pi_\theta$
  \end{algorithmic}
  \label{alg:iter-grpo}
\end{algorithm}

\paragraph{$\mathsf{KL}-$ Regularized RL \& On-Policy / Off-Policy Algorithms}
Finally putting together our clipped surrogate objective with the $\mathsf{KL}$ regularizer we obtain our final objective:
\begin{equation}
 \mathbb{E}_{x\sim \rho_{\mathcal{X}}}\mathcal{L}^c_{\alpha}(\pi(\cdot|x))-\beta \mathsf{KL}(\pi||\pi_{\mathrm{ref}}).
 \label{eq:KLLC}
 \end{equation}

We present the GRPO algorithm in Algorithm \ref{alg:iter-grpo} and the configurations that allow toggling between on-policy and off-policy GRPO in Table \ref{tab:configs}. Within the RL loop, the model is served for inference using vLLM \citep{kwon2023efficient}. The parameter $v$ controls how often the model is updated on the vLLM server (which corresponds to off-policy with $\alpha = \pi_{k-v+1}$). The parameter $i$ controls how many SGD iterations are applied to each batch sampled from the policy. For $v=1$ and $i=1$, the model is continuously served, and each batch of samples is used once in SGD. This corresponds to on-policy GRPO. For $i > 1$ and $v = 1$, the model is still continuously served, but each batch is used $i$ times in the SGD loop; this corresponds to an “off-policy” GRPO variant, as proposed in \cite{shao2024deepseekmath}. For large models that require tensor parallelism and multi-GPU serving, continuous model serving incurs additional communication costs. Our off-policy GRPO mitigates these costs by serving the model every $v > 1$ iterations (line 8 in Algorithm \ref{alg:iter-grpo}) and fixing $i = 1$. Our theory guarantees reward improvement as long as $v$ is not too large.\\

\paragraph{Computational and Communication Costs} Updating the model served by vLLM during GRPO training incurs varying costs depending on the model size, update frequency ($v$), and parallelism settings. When the training model and vLLM instance reside on different GPUs, or when vLLM uses tensor parallelism (TP), model updates may trigger deep copies and inter-GPU communication. These involve either full weight transfers or partitioned broadcasts, which scale linearly with model size. Frequent updates (e.g., $v=1$) can dominate the runtime, especially for large models (see the recent benchmark \cite{pytorch_vllm_benchmark} for latencies in serving large models with tensor parallelism using vLLM). To mitigate this, we update the vLLM model every $v>1$ iterations. This amortizes the copy cost while maintaining reward improvement guarantees from our theory. In our experiments (Section~\ref{sec:experiments}), we are limited to single node setups with relatively small models, and therefore cannot fully demonstrate the potential speedups —particularly those that would become more pronounced at larger scales. In our setups the speedups are modest, given that there is no inter GPU or inter nodes communication for serving the models.  See Section \ref{sec:limitation} for further discussion. \\

\noindent \textbf{On-Policy Clipped Objective with Zero Variance Masking a la DAPO \citep{yu2025dapoopensourcellmreinforcement}}
As discussed earlier in the interpretation of the lower bound in page 4, the samples with zero variance may lead to total variation terms to dominate the lower bound, hence we propose similar to DAPO \citep{yu2025dapoopensourcellmreinforcement} to mask these samples. For instance in the on policy case this would be with the following masked objective: 
\begin{equation}
 \mathbb{E}_{x\sim \rho_{\mathcal{X}}} \mathbbm{1}_{\sigma_{\pi_{k},r}(x)\neq 0}\left(\mathcal{L}^c_{\pi_{k}}(\pi(\cdot|x))-\beta \mathsf{KL}(\pi||\pi_{\mathrm{ref}})\right).
 \label{eq:KLLCZero}
 \end{equation}

\section{Related Work}\label{sec:relatedwork}
~~\\
\paragraph{Proximal Policy Optimization (PPO) and Extensions}
Proximal Policy Optimization (PPO) is a widely used on-policy reinforcement learning algorithm that improves training stability through clipped surrogate objectives. While PPO is effective in diverse settings, it is inherently limited by its on-policy nature, which constrains sample efficiency.
To address these limitations, several off-policy adaptations and extensions of PPO have been proposed. Generalized Proximal Policy Optimization (G-PPO) \citep{queeney2021generalized} enables sample reuse while maintaining convergence guarantees.  Transductive off-Policy PPO (ToPPO) \citep{gan2024transductive} builds on G-PPO  by incorporating transductive learning principles, bridging the gap between off-policy learning and theoretical guarantees of on-policy methods. Off-Policy PPO (OPPO) \citep{meng2023off} proposes novel corrections to integrate replay buffer samples in PPO-style updates.\\
\vskip -10 in
\paragraph{On-Policy and Off-Policy Actor-Critic Methods}
Actor-critic methods blend the strengths of policy gradients and value function estimation. Off-policy variants aim to improve sample efficiency by learning from a replay buffer. The Off-Policy Actor-Critic algorithm \citep{degris2012off} introduces importance weighting to enable stable updates from off-policy data. ACER \citep{wang2016sample} extends this with trust-region optimization and truncated importance sampling, enhancing by that the learning efficiency in discrete action spaces.
Mixing on-policy and off-policy methods aims to leverage the stability of on-policy updates with the efficiency of off-policy learning. P3O \citep{fakoor2020p3o} provides a principled approach that interleaves policy updates from both on- and off-policy data. \\

\paragraph{Off-Policy RLHF and other variants of GRPO} \citet{noukhovitch2025faster} introduced within the iterative DPO framework an asynchronous RLHF using off-policy data and that ensures faster convergence to the optimal policy.  New variants of GRPO have been proposed recently such as DAPO \citep{yu2025dapoopensourcellmreinforcement}
and DR-GRPO \citep{liu2025understandingr1zeroliketrainingcritical}. DAPO proposes the zero variance masking without theoretical backing, our work roots this in the improvement lower bound. DR-GRPO proposes to center only the reward without using the variance normalization. 
\section{Experiments}\label{sec:experiments}

\subsection{Ablation Studies on GSM8K}
~~\\
\paragraph{Setup, Model, and Data} In our first set of  experiments, we use  \href{https://huggingface.co/datasets/openai/gsm8k}{\texttt{GSM8K}} dataset from \cite{Cobbe2021} (MIT license), and {\texttt{Qwen/Qwen2.5-0.5B-Instruct}} (Apache 2.0 license) by \cite{Yang2024d}. We integrate our changes in Algorithm \ref{alg:iter-grpo} to the GRPO implementation in TRL \citep{vonwerra2022trl}, and train our models on  the training split of \texttt{GSM8K} on a node with 8 GPUs (GPU$_0$ for the vLLM server and 7 other GPUs for distributed training). See Appendix \ref{app:assets} for the hardware specification. We use a learning $5\times10^{-6}$ for all experiments and the KL regularizer $\beta=0.1$ in Equation \eqref{eq:KLLC}. 
We use the correctness of the LLM output as a reward. For GRPO training, the hyperparameters are the following: group size $G=16$ and per-device batch size $16$ (meaning each GPU processes a single prompt $x$ with $16$ responses). To increase the overall batchsize we use gradient accumulation of $4$, ending with an effective batch size of prompts of $28$. 
The context length used for this experiment is $200$, and the sampling temperature is  set  to $\tau=0.1$.\\

\begin{figure}[ht!]
  \centering

  \begin{subfigure}[t]{0.48\textwidth}
\includegraphics[width=\textwidth,keepaspectratio]{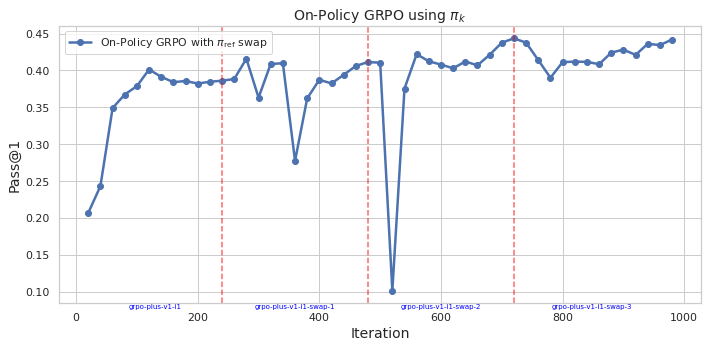}
    \caption{On-Policy GRPO with $\pi_{\mathrm{ref}}$ swap at end of each epoch. ($v=1$, $i=1$, $S=3$)}
\label{subfig:a}
  \end{subfigure}
  \hfill
  \begin{subfigure}[t]{0.48\textwidth}
    \includegraphics[width=\textwidth,keepaspectratio]{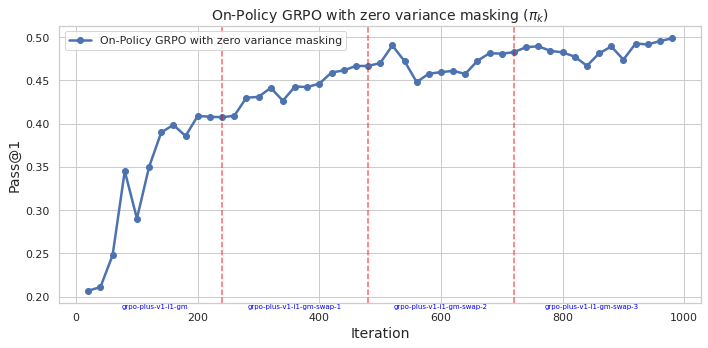}
    \caption{On-Policy GRPO with masking of samples with variance $\sigma_{\pi_{k},r}=0$, and with $\pi_{\mathrm{ref}}$ swap at end of each epoch. $v=1$, $i=1$, $S=3$)}
    \label{subfig:b}
  \end{subfigure}
\hfill
  \begin{subfigure}[t]{0.48\textwidth}
    \includegraphics[width=\textwidth,keepaspectratio]{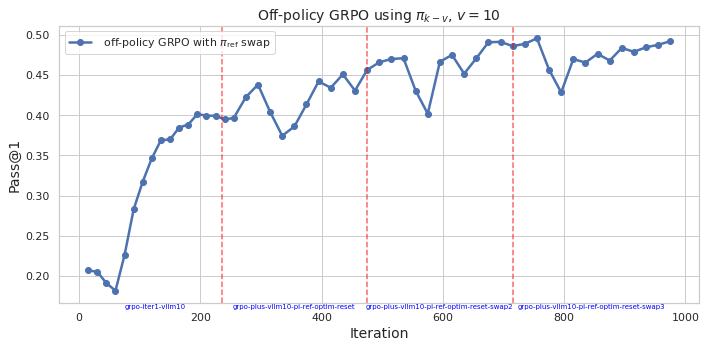}
    \caption{Off-Policy GRPO using $v=10$ (this amounts to fixing the model on the vLLM server for $10$ iterations and getting fresh samples for new batches), and with $\pi_{\mathrm{ref}}$ swap.($v=10$, $i=1$, $S=3$) }
     \label{subfig:c}
  \end{subfigure}
  \hfill
  \begin{subfigure}[t]{0.48\textwidth}
 \includegraphics[width=\textwidth,keepaspectratio]{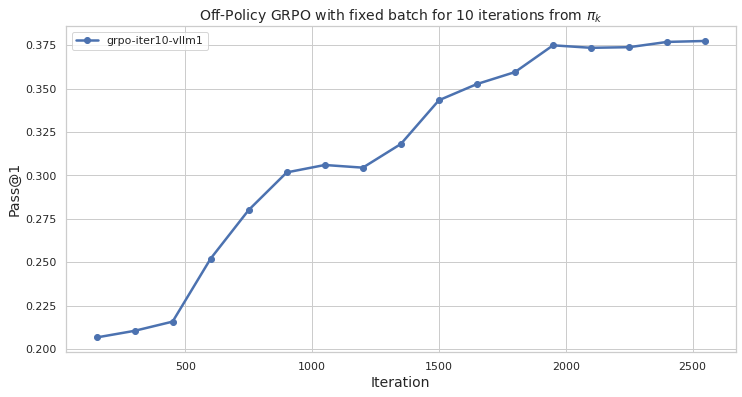}
    \caption{Off-Policy GRPO using fixed samples from $\pi_{k}$ for $10$ iterations. This will make $1$ epoch $10\times$ slower. $v=1$, $i=10$, $S=1$)  }
     \label{subfig:d}
    \vspace*{-1.2em}
  \end{subfigure}

  \caption{We train different variants of GRPO on the train portion of  GSM8K and report the Pass@1 on GSM8 test set using $50$ samples for each question in the test set for various variant of on-policy and off-policy GRPO. We see that as predicted by our theory, masking samples with zero variance stabilizes the training for on-policy training and leads to better performance. For off-policy training we see that using $v=10,i=1$ stabilizes also the training and leads also to better performance.}
  \label{fig:all-6}
\end{figure}

\paragraph{Ablations and Results} We train our models with GRPO using Algorithm \ref{alg:iter-grpo} with  a verifiable reward for answer correctness.  We  use for GRPO different configurations given in Table \ref{tab:configs} and report on the test split of \texttt{GSM8K} Pass@1 using 50 samples (i.e.\ frequency of success given 50 generations for each question) using the same sampling configuration as in training. We report results in Figure \ref{fig:all-6}: Fig.~\ref{subfig:a} for on-policy GRPO  ($i=1,v=1$) with the objective given in Equation \eqref{eq:KLLC} with $S=3$ (i.e.\ for 4 epochs with $\pi_{\mathrm{ref}}$ swap at end of each epoch with latest model); Fig.~\ref{subfig:b} for on-policy GRPO ($i=1,v=1$) with masking zero variance samples i.e.\ using the objective given Equation \eqref{eq:KLLCZero} with $S=3$;  Fig.~\ref{subfig:c} for our off-policy GRPO $(v=10,i=1)$, with $S=3$ and Fig.~\ref{subfig:d}  for \citet{shao2024deepseekmath}'s off-policy GRPO i.e $(v=1,i=10)$ for a single epoch. We see in Fig.~\ref{subfig:a} that while the on-policy GRPO converges to a maximum Pass@1 of $45\%$ it is unstable. The masking of zero variance sampling in \ref{subfig:b} stabilizes the on-policy GRPO and leads to an improvement of the performance to  $50\%$. This is in line with our theoretical grounding through the improvement lower bound. Our off-policy GRPO in Fig.~\ref{subfig:c} stabilizes the training also and leads to an improved Pass@1 of $50\%$ on the test set. In all three cases, we see that by resetting the $\pi_{\mathrm{ref}}$ to the latest model, GRPO amplifies the success rate above the current $\pi_{\mathrm{ref}}$, this concurs with the theoretical findings in \cite{mroueh2025reinforcementlearningverifiablerewards}.
Finally, the off-policy variant in \cite{shao2024deepseekmath} in Fig.~\ref{subfig:d} shows a slower convergence over an epoch.   \\

\subsection{Finetuning Qwen Distill R1 model (1.5 B) on Deepscaler Data }

In this section we use GRPO to finetune \href{https://huggingface.co/deepseek-ai/DeepSeek-R1-Distill-Qwen-1.5B}{\texttt{DeepSeek-R1-Distill-Qwen-1.5B}} \citep{guo2025deepseekR1} on  \href{https://huggingface.co/datasets/agentica-org/DeepScaleR-Preview-Dataset}{\texttt{DeepScaleR-Preview-Dataset}} from \cite{deepscaler2025} consisting of roughly  $40K$ math questions with known answers. We used \href{https://github.com/huggingface/Math-Verify}{\texttt{math-verify}} as the verifiable reward.  We use a learning rate of $1\times10^{-6}$ in the same distributed setting as before (GPU$_0$ for vLLM and 7 GPUs for distributed training). We use a  context length of $4096$, a group size $G=16$, a per-device batch size of $16$, and the KL regularizer is $\beta=0.001$. The sampling temperature used is $0.7$. We compared here the on-policy GRPO ($v=1,i=1$) to our off-policy GRPO ($v=10,i=1$) and report the performance of the trained model on a single epoch (around 24 hours on a single node). We report in Tables \ref{tab:aime24} and \ref{table:math500} Aime24 and Math500 performance using Huggingface light-eval \citep{lighteval}. Aime24 is evaluated with Pass@1 using 32 samples, and math500 with extractive matching as recommended in light-eval with a context length of $~32K$ (evaluation context length and all other sampling hyperparameters are set to the default in OpenR1 for this model). Plots of evaluation as function of iterations are given in Appendix \ref{app:plots}. We see that both on-policy and off-policy GRPO improve the performance of   \texttt{DeepSeek-R1-Distill-Qwen-1.5B} that has an Aime24 of $29\%$ to $32\%$ at maximum (over iterations), and its math-500 from $83\%$ to $87\%$. This result confirms our theoretical results that by going off-policy we don't loose in term of overall performance.

\begin{table}[ht!]
\centering
\begin{tabular}{lrrrr}
\toprule
       Model/Aime24 &     Min &     Max &  Median &    Mean \\
\midrule
  v1-i1-length-4096 &  0.2802 &  0.3229 &  0.3021 &  0.3022 \\
 v10-i1-length-4096 &  0.2781 &  0.3250 &  0.3047 &  0.3049 \\
\bottomrule
\end{tabular}
\caption{Aime24 using lighteval with on \& off-policy ( \texttt{(v1-i1)} and \texttt{(v10-i1)}) GRPO. }
\label{table:math500}
\end{table}
\begin{table}[ht!]
\centering
  \begin{tabular}{lrrrr}
\toprule
      Model/Math500 &    Min &    Max &  Median &    Mean \\
\midrule
  v1-i1-length-4096 &  0.830 &  0.870 &   0.854 &  0.8519 \\
 v10-i1-length-4096 &  0.822 &  0.872 &   0.846 &  0.8474 \\
\bottomrule
\end{tabular} 
\caption{Math 500 extractive matching using light-eval \citep{lighteval} with on and off-policy \texttt{(v1-i1)} and \texttt{(v10-i1)} GRPO. }
\label{tab:aime24}
\end{table}

\section{Conclusion and Discussion}

We revisited (on-policy) GRPO \citep{shao2024deepseekmath} and  showed that its clipping objective can be derived from first principles as a lower bound for reward improvement. We also gave theoretical  grounding to masking of zero variance samples suggested in DAPO \citep{yu2025dapoopensourcellmreinforcement}. We introduced off-policy GRPO and layed conditions under which it leads to policy improvement. Our off-policy GRPO has the advantage of reducing communication costs in serving the model for inference within the GRPO loop at each iteration as done in the on-policy counter-part, while not sacrificing performance. We showcased that off-policy GRPO stabilizes training and leads to either on par or improved performance as the on-policy one. 

The main takeaways of our paper to practitioners are: 
(1) Zero variance masking stabilizes on-policy GRPO's training 
(2) Off-policy GRPO attains its full potential in terms of maintaining  performance and lowering latencies and communication overhead in larger scale training where models are served using tensor parallelism (see \cite{pytorch_vllm_benchmark}). 

We hope our proof of concept for off-policy GRPO will help enabling stable and efficient reinforcement learning at scale.

\bibliographystyle{abbrvnat}
\bibliography{iclr2024_conference,rlhf} 

\begin{thebibliography}{32}
\providecommand{\natexlab}[1]{#1}
\providecommand{\url}[1]{\texttt{#1}}
\expandafter\ifx\csname urlstyle\endcsname\relax
  \providecommand{\doi}[1]{doi: #1}\else
  \providecommand{\doi}{doi: \begingroup \urlstyle{rm}\Url}\fi

\bibitem[Bai et~al.(2022)Bai, Jones, Ndousse, Askell, Chen, DasSarma, Drain,
  Fort, Ganguli, Henighan, et~al.]{bai2022training}
Y.~Bai, A.~Jones, K.~Ndousse, A.~Askell, A.~Chen, N.~DasSarma, D.~Drain,
  S.~Fort, D.~Ganguli, T.~Henighan, et~al.
\newblock Training a helpful and harmless assistant with reinforcement learning
  from human feedback.
\newblock \emph{arXiv preprint arXiv:2204.05862}, 2022.

\bibitem[Christiano et~al.(2017)Christiano, Leike, Brown, Martic, Legg, and
  Amodei]{NIPS2017_d5e2c0ad}
P.~F. Christiano, J.~Leike, T.~Brown, M.~Martic, S.~Legg, and D.~Amodei.
\newblock Deep reinforcement learning from human preferences.
\newblock In I.~Guyon, U.~V. Luxburg, S.~Bengio, H.~Wallach, R.~Fergus,
  S.~Vishwanathan, and R.~Garnett, editors, \emph{Advances in Neural
  Information Processing Systems}, volume~30. Curran Associates, Inc., 2017.
\newblock URL
  \url{https://proceedings.neurips.cc/paper_files/paper/2017/file/d5e2c0adad503c91f91df240d0cd4e49-Paper.pdf}.

\bibitem[Cobbe et~al.(2021)Cobbe, Kosaraju, Bavarian, Chen, Jun, Kaiser,
  Plappert, Tworek, Hilton, Nakano, Hesse, and Schulman]{Cobbe2021}
K.~Cobbe, V.~Kosaraju, M.~Bavarian, M.~Chen, H.~Jun, L.~Kaiser, M.~Plappert,
  J.~Tworek, J.~Hilton, R.~Nakano, C.~Hesse, and J.~Schulman.
\newblock Training verifiers to solve math word problems.
\newblock \emph{arXiv preprint arXiv:2110.14168}, 2021.

\bibitem[Degris et~al.(2012)Degris, White, and Sutton]{degris2012off}
T.~Degris, M.~White, and R.~S. Sutton.
\newblock Off-policy actor-critic.
\newblock \emph{arXiv preprint arXiv:1205.4839}, 2012.

\bibitem[Fakoor et~al.(2020)Fakoor, Chaudhari, and Smola]{fakoor2020p3o}
R.~Fakoor, P.~Chaudhari, and A.~J. Smola.
\newblock P3o: Policy-on policy-off policy optimization.
\newblock In \emph{Proceedings of The 35th Uncertainty in Artificial
  Intelligence Conference}, pages 1017--1027. PMLR, 2020.

\bibitem[Gan et~al.(2024)Gan, Yan, Tan, Wu, and Xing]{gan2024transductive}
Y.~Gan, R.~Yan, X.~Tan, Z.~Wu, and J.~Xing.
\newblock Transductive off-policy proximal policy optimization.
\newblock \emph{arXiv preprint arXiv:2406.03894}, 2024.

\bibitem[Guo et~al.(2025)Guo, Yang, Zhang, Song, Zhang, Xu, Zhu, Ma, Wang, Bi,
  et~al.]{guo2025deepseekR1}
D.~Guo, D.~Yang, H.~Zhang, J.~Song, R.~Zhang, R.~Xu, Q.~Zhu, S.~Ma, P.~Wang,
  X.~Bi, et~al.
\newblock Deepseek-r1: Incentivizing reasoning capability in llms via
  reinforcement learning.
\newblock \emph{arXiv preprint arXiv:2501.12948}, 2025.

\bibitem[Habib et~al.(2023)Habib, Fourrier, Kydlíček, Wolf, and
  Tunstall]{lighteval}
N.~Habib, C.~Fourrier, H.~Kydlíček, T.~Wolf, and L.~Tunstall.
\newblock Lighteval: A lightweight framework for llm evaluation, 2023.
\newblock URL \url{https://github.com/huggingface/lighteval}.

\bibitem[HuggingFace(2025{\natexlab{a}})]{Openr12025}
HuggingFace.
\newblock Open r1: A fully open reproduction of deepseek-r1, January
  2025{\natexlab{a}}.
\newblock URL \url{https://github.com/huggingface/open-r1}.

\bibitem[HuggingFace(2025{\natexlab{b}})]{huggingface_openr1_update3}
HuggingFace.
\newblock Open r1: Update \#3, Mar. 2025{\natexlab{b}}.
\newblock URL \url{https://huggingface.co/blog/open-r1/update-3}.
\newblock Accessed: 2025-05-11.

\bibitem[Kwon et~al.(2023)Kwon, Li, Zhuang, Sheng, Zheng, Yu, Gonzalez, Zhang,
  and Stoica]{kwon2023efficient}
W.~Kwon, Z.~Li, S.~Zhuang, Y.~Sheng, L.~Zheng, C.~H. Yu, J.~E. Gonzalez,
  H.~Zhang, and I.~Stoica.
\newblock Efficient memory management for large language model serving with
  pagedattention.
\newblock In \emph{Proceedings of the ACM SIGOPS 29th Symposium on Operating
  Systems Principles}, 2023.

\bibitem[Lambert et~al.(2024)Lambert, Morrison, Pyatkin, Huang, Ivison,
  Brahman, Miranda, Liu, Dziri, Lyu, et~al.]{lambert2024t}
N.~Lambert, J.~Morrison, V.~Pyatkin, S.~Huang, H.~Ivison, F.~Brahman, L.~J.~V.
  Miranda, A.~Liu, N.~Dziri, S.~Lyu, et~al.
\newblock T{\"u}lu 3: Pushing frontiers in open language model post-training.
\newblock \emph{arXiv preprint arXiv:2411.15124}, 2024.

\bibitem[Liu et~al.(2025)Liu, Chen, Li, Qi, Pang, Du, Lee, and
  Lin]{liu2025understandingr1zeroliketrainingcritical}
Z.~Liu, C.~Chen, W.~Li, P.~Qi, T.~Pang, C.~Du, W.~S. Lee, and M.~Lin.
\newblock Understanding r1-zero-like training: A critical perspective, 2025.
\newblock URL \url{https://arxiv.org/abs/2503.20783}.

\bibitem[Luo et~al.(2025)Luo, Tan, Wong, Shi, Tang, Roongta, Cai, Luo, Zhang,
  Li, Popa, and Stoica]{deepscaler2025}
M.~Luo, S.~Tan, J.~Wong, X.~Shi, W.~Y. Tang, M.~Roongta, C.~Cai, J.~Luo,
  T.~Zhang, L.~E. Li, R.~A. Popa, and I.~Stoica.
\newblock Deepscaler: Surpassing o1-preview with a 1.5b model by scaling rl.
\newblock \url{https://tinyurl.com/5e9rs33z}, 2025.
\newblock Notion Blog.

\bibitem[Meng et~al.(2023)Meng, Zheng, Pan, and Yin]{meng2023off}
W.~Meng, Q.~Zheng, G.~Pan, and Y.~Yin.
\newblock Off-policy proximal policy optimization.
\newblock \emph{Proceedings of the AAAI Conference on Artificial Intelligence},
  37\penalty0 (8):\penalty0 9162--9170, 2023.

\bibitem[Mroueh(2025)]{mroueh2025reinforcementlearningverifiablerewards}
Y.~Mroueh.
\newblock Reinforcement learning with verifiable rewards: Grpo's effective
  loss, dynamics, and success amplification, 2025.
\newblock URL \url{https://arxiv.org/abs/2503.06639}.

\bibitem[Noukhovitch et~al.(2025)Noukhovitch, Huang, Xhonneux, Hosseini,
  Agarwal, and Courville]{noukhovitch2025faster}
M.~Noukhovitch, S.~Huang, S.~Xhonneux, A.~Hosseini, R.~Agarwal, and
  A.~Courville.
\newblock Faster, more efficient {RLHF} through off-policy asynchronous
  learning.
\newblock In \emph{The Thirteenth International Conference on Learning
  Representations}, 2025.
\newblock URL \url{https://openreview.net/forum?id=FhTAG591Ve}.

\bibitem[Ouyang et~al.(2022)Ouyang, Wu, Jiang, Almeida, Wainwright, Mishkin,
  Zhang, Agarwal, Slama, Ray, et~al.]{ouyang2022training}
L.~Ouyang, J.~Wu, X.~Jiang, D.~Almeida, C.~Wainwright, P.~Mishkin, C.~Zhang,
  S.~Agarwal, K.~Slama, A.~Ray, et~al.
\newblock Training language models to follow instructions with human feedback.
\newblock \emph{Advances in Neural Information Processing Systems},
  35:\penalty0 27730--27744, 2022.

\bibitem[Paszke et~al.(2019)Paszke, Gross, Massa, Lerer, Bradbury, Chanan,
  Killeen, Lin, Gimelshein, Antiga, Desmaison, K{\"o}pf, Yang, DeVito, Raison,
  Tejani, Chilamkurthy, Steiner, Fang, Bai, and Chintala]{Paszke2019}
A.~Paszke, S.~Gross, F.~Massa, A.~Lerer, J.~Bradbury, G.~Chanan, T.~Killeen,
  Z.~Lin, N.~Gimelshein, L.~Antiga, A.~Desmaison, A.~K{\"o}pf, E.~Yang,
  Z.~DeVito, M.~Raison, A.~Tejani, S.~Chilamkurthy, B.~Steiner, L.~Fang,
  J.~Bai, and S.~Chintala.
\newblock {{PyTorch}}: {{An Imperative Style}}, {{High-Performance Deep
  Learning Library}}, Dec. 2019.

\bibitem[Queeney et~al.(2021)Queeney, Paschalidis, and
  Cassandras]{queeney2021generalized}
J.~Queeney, I.~C. Paschalidis, and C.~G. Cassandras.
\newblock Generalized proximal policy optimization with sample reuse.
\newblock In \emph{Advances in Neural Information Processing Systems},
  volume~34, 2021.

\bibitem[Schulman et~al.(2015)Schulman, Levine, Abbeel, Jordan, and
  Moritz]{schulman2015trust}
J.~Schulman, S.~Levine, P.~Abbeel, M.~Jordan, and P.~Moritz.
\newblock Trust region policy optimization.
\newblock In \emph{International conference on machine learning}, pages
  1889--1897. PMLR, 2015.

\bibitem[Schulman et~al.(2017)Schulman, Wolski, Dhariwal, Radford, and
  Klimov]{schulman2017proximal}
J.~Schulman, F.~Wolski, P.~Dhariwal, A.~Radford, and O.~Klimov.
\newblock Proximal policy optimization algorithms.
\newblock \emph{arXiv preprint arXiv:1707.06347}, 2017.

\bibitem[Shao et~al.(2024)Shao, Wang, Zhu, Xu, Song, Bi, Zhang, Zhang, Li, Wu,
  et~al.]{shao2024deepseekmath}
Z.~Shao, P.~Wang, Q.~Zhu, R.~Xu, J.~Song, X.~Bi, H.~Zhang, M.~Zhang, Y.~Li,
  Y.~Wu, et~al.
\newblock Deepseekmath: Pushing the limits of mathematical reasoning in open
  language models.
\newblock \emph{arXiv preprint arXiv:2402.03300}, 2024.

\bibitem[Stiennon et~al.(2020)Stiennon, Ouyang, Wu, Ziegler, Lowe, Voss,
  Radford, Amodei, and Christiano]{stiennon2020learning}
N.~Stiennon, L.~Ouyang, J.~Wu, D.~Ziegler, R.~Lowe, C.~Voss, A.~Radford,
  D.~Amodei, and P.~F. Christiano.
\newblock Learning to summarize with human feedback.
\newblock \emph{Advances in Neural Information Processing Systems},
  33:\penalty0 3008--3021, 2020.

\bibitem[vLLM(2025)]{pytorch_vllm_benchmark}
P.~vLLM.
\newblock Pytorch ci hud: vllm benchmark dashboard.
\newblock
  \url{https://hud.pytorch.org/benchmark/llms?repoName=vllm-project/vllm},
  2025.
\newblock Accessed: 2025-05-14.

\bibitem[Vojnovic and Yun(2025)]{vojnovic2025alignmentobjectivegrpo}
M.~Vojnovic and S.-Y. Yun.
\newblock What is the alignment objective of grpo?, 2025.
\newblock URL \url{https://arxiv.org/abs/2502.18548}.

\bibitem[von Werra et~al.(2020{\natexlab{a}})von Werra, Belkada, Tunstall,
  Beeching, Thrush, Lambert, Huang, Rasul, and Gallouédec]{Vonwerra2022}
L.~von Werra, Y.~Belkada, L.~Tunstall, E.~Beeching, T.~Thrush, N.~Lambert,
  S.~Huang, K.~Rasul, and Q.~Gallouédec.
\newblock Trl: Transformer reinforcement learning.
\newblock \url{https://github.com/huggingface/trl}, 2020{\natexlab{a}}.

\bibitem[von Werra et~al.(2020{\natexlab{b}})von Werra, Belkada, Tunstall,
  Beeching, Thrush, Lambert, Huang, Rasul, and Gallouédec]{vonwerra2022trl}
L.~von Werra, Y.~Belkada, L.~Tunstall, E.~Beeching, T.~Thrush, N.~Lambert,
  S.~Huang, K.~Rasul, and Q.~Gallouédec.
\newblock Trl: Transformer reinforcement learning.
\newblock \url{https://github.com/huggingface/trl}, 2020{\natexlab{b}}.

\bibitem[Wang et~al.(2016)Wang, Bapst, Heess, Mnih, Munos, Kavukcuoglu, and
  De~Freitas]{wang2016sample}
Z.~Wang, V.~Bapst, N.~Heess, V.~Mnih, R.~Munos, K.~Kavukcuoglu, and
  N.~De~Freitas.
\newblock Sample efficient actor-critic with experience replay.
\newblock \emph{arXiv preprint arXiv:1611.01224}, 2016.

\bibitem[Wolf et~al.(2020)Wolf, Debut, Sanh, Chaumond, Delangue, Moi, Cistac,
  Rault, Louf, Funtowicz, Davison, Shleifer, von Platen, Ma, Jernite, Plu, Xu,
  Scao, Gugger, Drame, Lhoest, and Rush]{Wolf2020}
T.~Wolf, L.~Debut, V.~Sanh, J.~Chaumond, C.~Delangue, A.~Moi, P.~Cistac,
  T.~Rault, R.~Louf, M.~Funtowicz, J.~Davison, S.~Shleifer, P.~von Platen,
  C.~Ma, Y.~Jernite, J.~Plu, C.~Xu, T.~L. Scao, S.~Gugger, M.~Drame, Q.~Lhoest,
  and A.~M. Rush.
\newblock {{HuggingFace}}'s {{Transformers}}: {{State-of-the-art Natural
  Language Processing}}, July 2020.

\bibitem[Yang et~al.(2024)Yang, Yang, Hui, Zheng, Yu, Zhou, Li, Li, Liu, Huang,
  Dong, Wei, Lin, Tang, Wang, Yang, Tu, Zhang, Ma, Yang, Xu, Zhou, Bai, He,
  Lin, Dang, Lu, Chen, Yang, Li, Xue, Ni, Zhang, Wang, Peng, Men, Gao, Lin,
  Wang, Bai, Tan, Zhu, Li, Liu, Ge, Deng, Zhou, Ren, Zhang, Wei, Ren, Liu, Fan,
  Yao, Zhang, Wan, Chu, Liu, Cui, Zhang, Guo, and Fan]{Yang2024d}
A.~Yang, B.~Yang, B.~Hui, B.~Zheng, B.~Yu, C.~Zhou, C.~Li, C.~Li, D.~Liu,
  F.~Huang, G.~Dong, H.~Wei, H.~Lin, J.~Tang, J.~Wang, J.~Yang, J.~Tu,
  J.~Zhang, J.~Ma, J.~Yang, J.~Xu, J.~Zhou, J.~Bai, J.~He, J.~Lin, K.~Dang,
  K.~Lu, K.~Chen, K.~Yang, M.~Li, M.~Xue, N.~Ni, P.~Zhang, P.~Wang, R.~Peng,
  R.~Men, R.~Gao, R.~Lin, S.~Wang, S.~Bai, S.~Tan, T.~Zhu, T.~Li, T.~Liu,
  W.~Ge, X.~Deng, X.~Zhou, X.~Ren, X.~Zhang, X.~Wei, X.~Ren, X.~Liu, Y.~Fan,
  Y.~Yao, Y.~Zhang, Y.~Wan, Y.~Chu, Y.~Liu, Z.~Cui, Z.~Zhang, Z.~Guo, and
  Z.~Fan.
\newblock Qwen2 {{Technical Report}}, Sept. 2024.

\bibitem[Yu et~al.(2025)Yu, Zhang, Zhu, Yuan, Zuo, Yue, Fan, Liu, Liu, Liu,
  Lin, Lin, Ma, Sheng, Tong, Zhang, Zhang, Zhang, Zhu, Zhu, Chen, Chen, Wang,
  Yu, Dai, Song, Wei, Zhou, Liu, Ma, Zhang, Yan, Qiao, Wu, and
  Wang]{yu2025dapoopensourcellmreinforcement}
Q.~Yu, Z.~Zhang, R.~Zhu, Y.~Yuan, X.~Zuo, Y.~Yue, T.~Fan, G.~Liu, L.~Liu,
  X.~Liu, H.~Lin, Z.~Lin, B.~Ma, G.~Sheng, Y.~Tong, C.~Zhang, M.~Zhang,
  W.~Zhang, H.~Zhu, J.~Zhu, J.~Chen, J.~Chen, C.~Wang, H.~Yu, W.~Dai, Y.~Song,
  X.~Wei, H.~Zhou, J.~Liu, W.-Y. Ma, Y.-Q. Zhang, L.~Yan, M.~Qiao, Y.~Wu, and
  M.~Wang.
\newblock Dapo: An open-source llm reinforcement learning system at scale,
  2025.
\newblock URL \url{https://arxiv.org/abs/2503.14476}.

\end{thebibliography}

\newpage
\appendix

\section{Broader Impact and Limitations}\label{sec:limitation}
Our work analyzes the celebrated GRPO algorithm and develops an adaptation for the off-policy setting motivated by recent efforts for PPO that demonstrated higher stability and efficiency.
Our primary contributions are theoretical, providing formal conditions under which advantage optimization guarantees  policy improvement for the on-policy  and  off-policy regimes.
These insights provide lower bounds  on policy improvement and  directly inform a practical clipped surrogate optimization objective for large language model (LLM) policy training that inherits our theoretical guarantees for both on policy and off policy regimes. In the on-policy regime our lower bound shed the light  and give theoretical backing to  the benefits of  masking samples with zero variance as suggested in the DAPO paper \citep{yu2025dapoopensourcellmreinforcement}.
Our formulation also clarifies  theoretical relationships between  our newly introduced off-policy GRPO, PPO variants, and general off-policy optimization frameworks -- a linkage previously underexplored in the literature.
Our derived off-policy GRPO algorithm is validated experimentally demonstrating improved performance compared to standard GRPO, while having the potential to  reduce the communication overhead across devices in serving large models for sampling that is  needed in GRPO.
The broader impacts that we anticipate from our work (beside those directly inherited from GRPO and reinforcement fine-tuning of LLMs and the risks associated to the dual use of the enabled reasoning models) are then generally positive, as it enhances RL efficiency, reducing computational costs and improving stability.

The main limitation of our work is that the empirical validation remains constrained to smaller datasets, smaller model architectures, and smaller context size (4096 tokens at maximum) that can be trained on our hardware setup consisting of one compute node with 8 H100 NVIDIA gpus (1 used for the vLLM server and 7 for training the policy LLM). Our 1.5 B experimental setup, with deepscaler data is at the limit of what can fit in the memory of a single node.

This limitation primarily reflects the common resource constraints associated with provisioning large-scale distributed training environments, rather than any inherent restriction of the algorithm itself. Note that for larger context, larger batch size and larger architectures than the ones used in our paper, multi-node training is required.   

While our main contribution here remains theoretical and backed with ablation studies on a single node, we reserve to scale up our experiments to larger training runs in future work aimed at showcasing the fact that the benefits of our off-policy algorithms in terms of efficient and reduced communication are expected to become even more pronounced in the large-scale distributed regime as it is already showed in multiple off policy RL works.

\section{Assets}\label{app:assets}

\paragraph{Hardware setup} All our experiments were run on one compute node with Dual 48-core Intel Xeon 8468, 2TB of RAM, 8 NVIDIA HGX H100 80GB SMX5, 8x 3.4TB Enterprise NVMe U.2 Gen4, and 10x NVIDIA Mellanox Infiniband Single port NDR adapters, running RedHat Enterprise Linux 9.5\\

\paragraph{Libraries} Our experiments rely on the open-source libraries \href{https://pytorch.org/}{\texttt{pytorch}} \citep{Paszke2019} (license: BSD), \href{https://github.com/huggingface/transformers}{\texttt{HuggingFace Transformers}} \citep{Wolf2020} (Apache 2.0 license), and \href{https://github.com/huggingface/trl}{\texttt{HuggingFace TRL}} \citep{Vonwerra2022} (Apache 2.0 license). We also relied on Open-R1 \citep{Openr12025} as well as light-eval  \citep{lighteval} for the evaluation of Aime24 and Math500.\\

\paragraph{Code re-use} Our GRPO training code is based on the public Github repository \url{https://github.com/huggingface/open-r1} \citep{Openr12025}.\\

\paragraph{Data and Models} In our experiments, we use following publicly available datasets: (1) \href{https://huggingface.co/datasets/openai/gsm8k}{\texttt{GSM8K}} dataset from \cite{Cobbe2021} (MIT license), and (2) the \href{https://huggingface.co/datasets/agentica-org/DeepScaleR-Preview-Dataset}{\texttt{DeepScaleR-Preview-Dataset}} from \cite{deepscaler2025} (MIT license).
The models that we used were \href{https://huggingface.co/Qwen/Qwen2.5-0.5B}{\texttt{Qwen/Qwen2.5-0.5B-Instruct}} (Apache 2.0 license) by \cite{Yang2024d}, and \href{https://huggingface.co/deepseek-ai/DeepSeek-R1-Distill-Qwen-1.5B}{\texttt{DeepSeek-R1-Distill-Qwen-1.5B}} (MIT license) by \cite{guo2025deepseekR1}.

\section{Reward Improvement Lower Bound}
\subsection{Proof of Theorem \ref{lem:offpolicyimprovement} }

We have :
\[J(\pi (\cdot| x)) = \mathbb{E}_{y \sim \pi(\cdot|x) } r(x,y)  \]

Let $\pi_{k}$ be the current policy and  $\alpha(\cdot|x)$ be another policy typically consider $\alpha(\cdot|x) = \pi_{k-i}(\cdot | x)$.

Define mean and variances of the off-policy reward, i.e policy under $\alpha$:

$ \mu_{\alpha} (x) =  \mathbb{E}_{y \sim \alpha(\cdot|x)} r(x,y)   $ and 
$ \sigma_{\alpha}(x) = \sqrt{ \mathbb{E}_{y \sim \alpha(\cdot|x)}( r(x,y) - \mu_{\alpha} (x))^2 },$
and denote for $0<\varepsilon<1$:
$ \sigma_{\alpha,\varepsilon}(x) =\sqrt{\sigma^2_{\alpha}(x) + \varepsilon}$.

Note that we have a bounded reward $0\leq r(x,y) \leq \nor{r}_{\infty}$ which implies that
$\sigma^2_{\alpha}(x) \leq \frac{ \nor{r}^2_{\infty}}{4}$, and hence we have:
\[ \sigma_{\alpha,\varepsilon}(x) \leq \sqrt{\frac{\nor{r}^2_{\infty}}{4} + \varepsilon}.  \]

We normalize the reward so that :
$\sigma_{\alpha,\varepsilon}(x) \leq \sqrt{\frac{\nor{r}^2_{\infty}}{4} + \varepsilon} \leq 1 . $

We denote GRPO advantage function as: 
\[ A_{\alpha} (x,y)   =  \frac{r(x,y) - \mu_{\alpha} (x)  }{\sigma_{\alpha,\varepsilon}(x)}\]
\[ \mathcal{L}_{\alpha}(\pi (\cdot | x)) = \mathbb{E}_{y \sim \alpha(\cdot|x)} \frac{\pi(y|x)}{\alpha(y|x)} A_{\alpha}(x,y)  \]

If $\alpha= \pi_{k}$, we obtain the online policy objective function of GRPO, where the advantage is computed with the current policy $\pi_{k}$, i.e using $A_{\pi_{k}}(x,y)$.

We have: 
\begin{align*}
\mathcal{L}_{\alpha}(\pi (\cdot | x)) &= \frac{1}{\sigma_{\alpha,\varepsilon}(x)} \left( \mathbb{E}_{y \sim \pi(\cdot |x )} r(x,y) - \mu_{\alpha} (x)\right)\\
&= \frac{1}{\sigma_{\alpha,\varepsilon}(x) } J(\pi(\cdot|x) ) - \frac{1}{\sigma_{\alpha,\epsilon}(x)} J(\alpha(\cdot|x)) 
\end{align*}

Our goal is to provide an upper bound on :

\[\mathcal{L}_{\alpha}(\pi (\cdot | x)) -   \left( J(\pi(\cdot|x )) - J( \pi_{k}(\cdot|x) ) \right) \]

Hence we have: 
\begin{align*}
&\mathcal{L}_{\alpha}(\pi (\cdot | x)) -   \left( J(\pi(\cdot|x )) - J( \pi_{k}(\cdot|x) ) \right) = \left(\frac{1}{\sigma_{\alpha,\varepsilon}(x)} -1 \right) J(\pi(\cdot |x)) +  J(\pi_{k}(\cdot|x) ) - \frac{1}{\sigma_{\alpha,\varepsilon}(x)} J(\alpha(\cdot|x))\\
&= \left(\frac{1}{\sigma_{\alpha,\varepsilon}(x)} -1 \right) \left(J(\pi(\cdot |x)) - J(\alpha(\cdot|x)) + J(\alpha(\cdot| x)) \right) +  J(\pi_{k}(\cdot|x) ) - \frac{1}{\sigma_{\alpha,\varepsilon}(x)} J(\alpha(\cdot|x))\\
& = \frac{1-\sigma_{\alpha,\varepsilon}(x)}{\sigma_{\alpha,\varepsilon}(x)} \left(J(\pi(\cdot|x)) - J (\alpha(\cdot|x))\right)+ ( J(\pi_{k}(\cdot|x)) - J(\alpha(\cdot|x))) +  \frac{1}{\sigma_{\alpha,\varepsilon}(x)} J(\alpha(\cdot|x))- \frac{1}{\sigma_{\alpha,\varepsilon}(x)} J(\alpha(\cdot|x))\\
& =  \frac{1-\sigma_{\alpha,\varepsilon}(x)}{\sigma_{\alpha,\varepsilon}(x)} \left(J(\pi(\cdot|x)) - J (\alpha(\cdot|x))\right)+ ( J(\pi_{k}(\cdot|x)) - J(\alpha(\cdot|x))) 
\end{align*}

\begin{lemma}[Kantorovich-Rubenstein duality of total variation distance, see ]\label{lem: tv} The Kantorovich-Rubenstein duality (variational representation) of the total variation distance is as follows:
\begin{equation}\label{Eq: tv rep}
    \mathbb{TV}(m_1,m_2)=\frac{1}{2L}\sup_{g \in \mathcal{G}_L}\left\{\mathbb{E}_{Z\sim m_1}[g(Z)]-\mathbb{E}_{Z\sim m_2}[g(Z)]\right\},
\end{equation}
where $\mathcal{G}_L=\{g: \mathcal{Z}\rightarrow \mathbb{R}, ||g||_\infty \leq L \}$. 
\end{lemma}

On the other hand using Lemma \ref{lem: tv} we have:

\[ J(\pi(\cdot|x)) - J (\alpha(\cdot|x)) \leq 2 \nor{r}_{\infty} \mathbb{TV}(\pi(\cdot|x),\alpha(\cdot|x)) \]
and 
\[   
J(\pi_{k}(\cdot|x)) - J(\alpha(\cdot|x)) \leq 2 \nor{r}_{\infty}  \mathbb{TV}(\pi_k(\cdot|x),\alpha(\cdot|x))
\]

By our assumption on the reward  we have :
\[ \frac{1- \sigma_{\alpha,\varepsilon}(x)}{\sigma_{\alpha,\varepsilon}(x)}  \geq 0\]
so that we obtain the final bound as follows:
\begin{align*}
\mathcal{L}_{\alpha}(\pi (\cdot | x)) -   \left( J(\pi(\cdot|x )) - J( \pi_{k}(\cdot|x) ) \right) \leq  2 \frac{1- \sigma_{\alpha,\varepsilon}(x)}{\sigma_{\alpha,\varepsilon}(x)} \nor{r}_{\infty} \mathbb{TV}(\pi(\cdot|x),\alpha(\cdot|x)) + 2 \nor{r}_{\infty} \mathbb{TV}(\pi_k(\cdot|x),\alpha(\cdot|x))
\end{align*}

We obtain finally our lower bound on policy improvement as follows:

\boxed{J(\pi(\cdot|x )) - J( \pi_{k}(\cdot|x) ) \geq \mathcal{L}_{\alpha}(\pi (\cdot | x)) - 2 \frac{1- \sigma_{\alpha,\varepsilon}(x)}{\sigma_{\alpha,\varepsilon}(x)} \nor{r}_{\infty} \mathbb{TV}(\pi(\cdot|x),\alpha(\cdot|x)) - 2 \nor{r}_{\infty} \mathbb{TV}(\pi_k(\cdot|x),\alpha(\cdot|x))}

Integrating over $x$ (the prompts) we have:
\begin{align*}
&\mathbb{E}_{x\sim \rho_{\mathcal{X}}}J(\pi(\cdot|x )) - \mathbb{E}_{x\sim \rho_{\mathcal{X}}}J( \pi_{k}(\cdot|x) ) \geq \mathbb{E}_{x\sim \rho_{\mathcal{X}}}\mathcal{L}_{\alpha}(\pi (\cdot | x)) - 2 \nor{r}_{\infty} \mathbb{E}_{x\sim \rho_{\mathcal{X}}}\frac{1- \sigma_{\alpha,\varepsilon}(x)}{\sigma_{\alpha,\varepsilon}(x)}  \mathbb{TV}(\pi(\cdot|x),\alpha(\cdot|x))\\
&- 2 \nor{r}_{\infty} \mathbb{E}_{x\sim \rho_{\mathcal{X}}} \mathbb{TV}(\pi_k(\cdot|x),\alpha(\cdot|x))\\
&\geq \mathbb{E}_{x\sim \rho_{\mathcal{X}}}\mathcal{L}_{\alpha}(\pi (\cdot | x)) - 2 \nor{r}_{\infty} \sqrt{\mathbb{E}_{x\sim \rho_{\mathcal{X}}}\frac{(1- \sigma_{\alpha,\varepsilon}(x))^2}{\sigma^2_{\alpha,\varepsilon}(x)} } \sqrt{ \mathbb{E}_{x\sim \rho_{\mathcal{X}}}\mathbb{TV}^2(\pi(\cdot|x),\alpha(\cdot|x))}- 2 \nor{r}_{\infty} \mathbb{E}_{x\sim \rho_{\mathcal{X}}} \mathbb{TV}(\pi_k(\cdot|x),\alpha(\cdot|x))
\end{align*}

\section{Experiments}\label{app:plots}

\begin{figure}[htbp]
  \centering

  \begin{subfigure}[t]{\textwidth}
    \includegraphics[width=\textwidth,keepaspectratio]{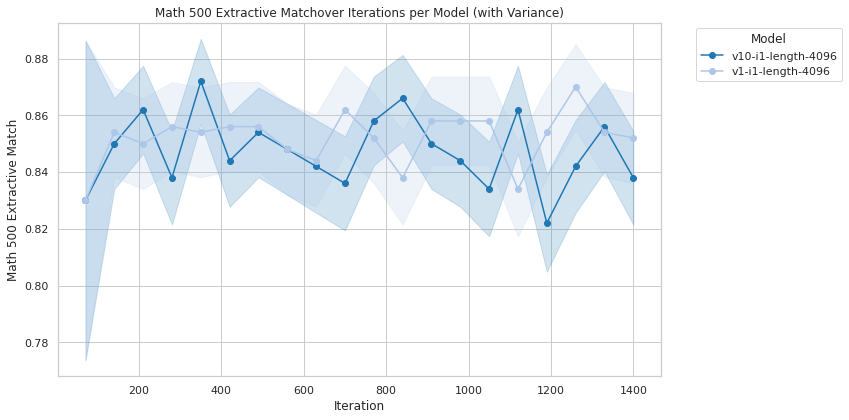}
    \caption{Aime 24}
  \end{subfigure}
  \hfill
  \begin{subfigure}[t]{\textwidth}
    \includegraphics[width=\textwidth,keepaspectratio]{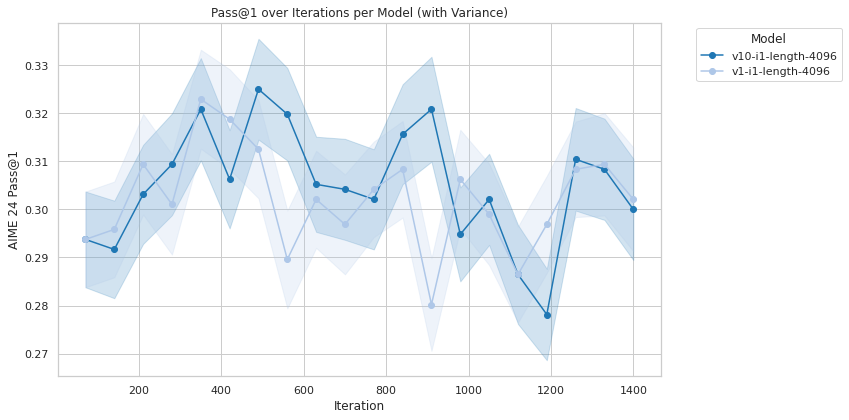}
    \caption{Math 500.}
  \end{subfigure}
  \caption{Aime 24/ Math 500}
  \end{figure}

\end{document}